\pdfoutput=1

\documentclass[11pt]{article}

\usepackage[final]{acl}

\usepackage{times}
\usepackage{latexsym}
\usepackage{amsmath, amsthm, amssymb, amsfonts}
\usepackage[T1]{fontenc}

\usepackage[utf8]{inputenc}

\usepackage[english]{babel} 

\usepackage{microtype}

\usepackage{inconsolata}

\usepackage{graphicx}
\usepackage{xcolor}
\usepackage{tikz}
\usepackage[many]{tcolorbox}
\usepackage{textcomp}
\usepackage{multirow}
\usepackage{wrapfig}
\usepackage{caption}
\usepackage{soul}
\usepackage{adjustbox}
\usepackage{enumitem}
\usepackage{caption}
\usepackage{tabularx}
\usepackage{listings}
\usepackage{dsfont}
\definecolor{CB_gray}{gray}{0.5}
\newcommand{\mycolon}{\ensuremath{\thinspace \colon \thinspace}}

\newcommand{\tuple}[1]{\left \langle #1\right\rangle}

\tcbset{prompt/.style={
    enhanced,
    size=fbox,
    boxrule=3pt,
    arc=2mm,
    auto outer arc,
    left=10pt,
    right=10pt,
    top=10pt,
    bottom=10pt,
    fontupper=\fontfamily{cmtt}\selectfont, 
    colback=CB_gray!10,
    colframe=CB_gray!25,
    coltitle=CB_gray!100, 
}}

\tcbset{vignette/.style={
    enhanced,
    size=fbox,
    boxrule=3pt,
    arc=2mm,
    auto outer arc,
    left=10pt,
    right=10pt,
    top=10pt,
    bottom=10pt,
    fontupper=\fontfamily{Avenir}\selectfont, 
    colback=CB_gray!10,
    colframe=CB_gray!25,
    coltitle=CB_gray!100, 
}}
\definecolor{codegreen}{rgb}{0,0.6,0}
\definecolor{codegray}{rgb}{0.5,0.5,0.5}
\definecolor{codepurple}{rgb}{0.58,0,0.82}
\definecolor{backcolour}{rgb}{0.95,0.95,0.92}

\lstdefinestyle{mystyle}{
  commentstyle=\color{codegreen},
  keywordstyle=\color{magenta},
  numberstyle=\tiny\color{codegray},
  basicstyle=\ttfamily\footnotesize,
  stringstyle=\color{codepurple},
  breakatwhitespace=false,         
  breaklines=true,                 
  captionpos=b,                    
  keepspaces=true,                 
  numbers=left,                    
  numbersep=5pt,                  
  showspaces=false,                
  showstringspaces=false,
  showtabs=false,                  
  tabsize=2
}
\definecolor{firebrick}{rgb}{0.7, 0.13, 0.13}
\definecolor{battleshipgrey}{rgb}{0.52, 0.52, 0.51}
\definecolor{goldenpoppy}{rgb}{0.99, 0.76, 0.0}
\definecolor{cerulean}{rgb}{0.0, 0.48, 0.65}

\title{Integrating Neural and Symbolic Components \\ in a Model of Pragmatic Question-Answering}

\author{Polina Tsvilodub \\
  University of T\"ubingen \\
  \texttt{first.last@uni-tuebingen.de} \\\And
 Robert D. Hawkins \\
  Stanford University \\
  \texttt{hawkrobe@gmail.com} \\ \And
  Michael Franke \\
  University of T\"ubingen \\
  \texttt{first.last@uni-tuebingen.de}}

\begin{document}
\maketitle
\begin{abstract}

Computational models of pragmatic language use have traditionally relied on hand-specified sets of utterances and meanings, limiting their applicability to real-world language use. 
We propose a neuro-symbolic framework that enhances probabilistic cognitive models by integrating LLM-based modules to propose and evaluate key components in natural language, eliminating the need for manual specification. 
Through a classic case study of pragmatic question-answering, we systematically examine various approaches to incorporating neural modules into the cognitive model—from evaluating utilities and literal semantics to generating alternative utterances and goals. 
We find that hybrid models can match or exceed the performance of traditional probabilistic models in predicting human answer patterns. 
However, the success of the neuro-symbolic model depends critically on how LLMs are integrated: while they are particularly effective for proposing alternatives and transforming abstract goals into utilities, they face challenges with truth-conditional semantic evaluation.
This work charts a path toward more flexible and scalable models of pragmatic language use while illuminating crucial design considerations for balancing neural and symbolic components.
\end{abstract}

\section{Introduction}
\label{sec:introduction}
Imagine you are a barista in a café with only three items in stock: iced coffee, soda, and Chardonnay. 
If a customer asks: ``Do you have iced tea?'', you might naturally respond ``I'm sorry, we don't have iced tea, but I can make you an iced coffee!''. 
This situation exemplifies \textit{pragmatic question answering}, where answerers commonly go beyond the literal question being asked \cite{clark1979responding}.
Classical accounts of the semantic meaning of questions and answers \citep[e.g.,][]{hamblin1973questions, groenendijk1984studies, hakulinen2001minimal}, maintain that polar questions like ``Do you have iced tea?'' are fully resolved by a polar answer \{yes, no\}.
Yet humans routinely provide a \textit{relevant} selection of additional information (e.g., mentioning the iced coffee, but not the Chardonnay). 

Understanding what, exactly, makes an answer relevant has been a central question in the field of pragmatics, with extensive work investigating the contextual factors that shape answer selection \citep[e.g.][]{van2003questioning,StevensBenz2016:Pragmatic-quest,rothe2017question}.
One recent framework for modeling these pragmatic choices is the Rational Speech Act framework \citep{frank2012predicting, degen2023rational}, which has been successfully applied to both question and answer selection \cite{Hawkins2015WhyDY,HawkinsGoodman2017:Why-Do-You-Ask-,hawkins2025relevant}.
The probabilistic cognitive models (PCMs) developed within this framework offer significant advantages through their transparent, explicit task decomposition and systematic error analysis \citep{farrell2018computational}. 

However, these models are typically limited to a small set of predefined examples, restricting their applicability to real-world scenarios. 
In contrast, Large Language Models (LLMs) offer a complementary set of capabilities.
They can process open-ended natural language input and generate flexible responses, but often struggle with subtle pragmatic patterns \citep{hu-etal-2023-fine,RuisKhan2023:The-Goldilocks-,TsvilodubMarty2024:Experimental-Pr} and
lack the degree of explainability that makes PCMs so valuable for cognitive modeling  \citep{zhao2023explainability}.

To address these complementary strengths and limitations, we explore a family of \textit{neuro-symbolic} models, with different combinations of both approaches to leverage their respective strengths and to overcome known shortcomings.\footnote{We use the term \textit{neuro-symbolic} in the sense of a model that has \textit{neural} network components (here, LLMs), that are scaffolded by a \textit{symbolic} task analysis, i.e., integrated in a particular computational procedure. Other senses of the term also exist \citep{bhuyan2024neuro}.}  
Our approach builds on the task analysis developed in previous work on pragmatic question-answering \citep{Hawkins2015WhyDY, HawkinsGoodman2017:Why-Do-You-Ask-, hawkins2025relevant} in two ways. First, we use it as a \textit{scaffolding structure} that determines the computational steps, with LLMs executing specific subtasks that would traditionally require manual specification in a PCM (Sections~\ref{sec:pcm-llm-evaluators}--\ref{sec:pcm-llm-proposers}).
Second, we verbalize (parts of) the scaffolding structure in a single prompt, relying on a single LLM call to solve the respective computational task (Section~\ref{sec:llm-scaffolding}).
This dual approach enables us to systematically investigate the tradeoffs between fine-grained task decomposition and end-to-end neural processing.

Our key contributions are as follows: 
\begin{itemize}
    \item A novel neuro-symbolic framework that extends probabilistic models of pragmatic question answering to more open-ended natural language.
    \item A systematic investigation of how different integrations of neural and symbolic components affect model behavior.
    
    \item Empirical validation against human data, demonstrating that neuro-symbolic models can match or exceed traditional probabilistic approaches in predicting human behavior.
\end{itemize}

\begin{figure}[t]
	\centering
	\includegraphics[width=0.48\textwidth]{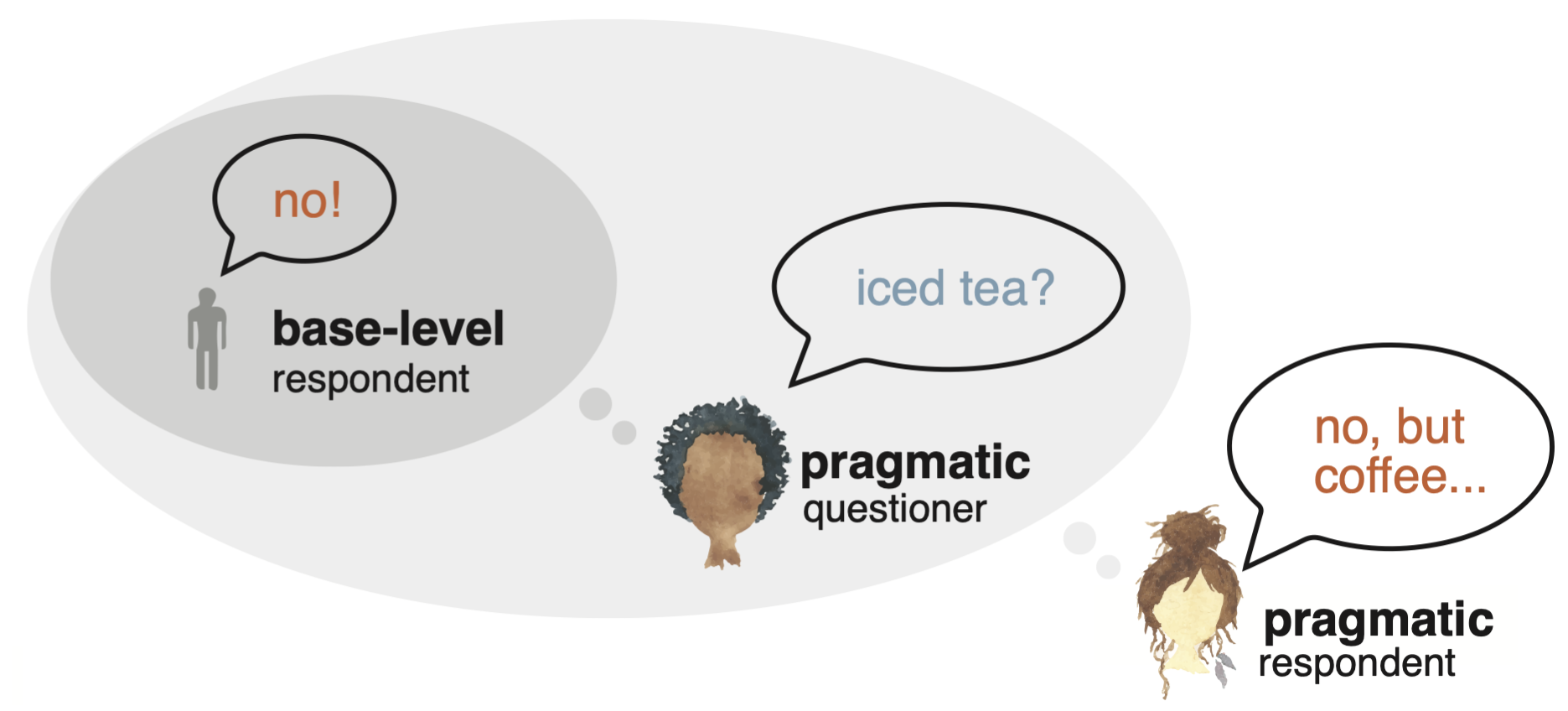}
	\caption{\label{fig:prior-pq} Probabilistic cognitive model (PCM) of pragmatic question answering. The PCM, built in the Rational Speech Act framework, implements recursive back-and-forth reasoning of rational agents. The questioner chooses a question based on their decision problem and an expectation of responses that any question might provoke. The respondent chooses a relevant response based on the decision problem inferred from the question.
    }
\end{figure}

\section{A Probabilistic Cognitive Model of Relevant Question-Answering}
\label{sec:prior-pq}


The probabilistic cognitive model we use for task analysis and scaffolding, which we refer to as the \textit{QA model} \citep{hawkins2025relevant}, captures a rational \textit{pragmatic respondent} that chooses an answer by reasoning about how a pragmatic questioner chooses a question (see Figure~\ref{fig:prior-pq} for overview and Appendix~\ref{sec:app:qa-model} for technical detail). 
The questioner is grounded in a context-independent \textit{base-level respondent}. 
The pragmatic questioner selects a question based on the response they expect to get from the base-level respondent, who answers austerely without considering the wider context.
The pragmatic respondent, in turn, reasons about the motivation of the speaker for asking the question (i.e., \textit{infers} their goal from the question) and chooses responses that are expected to be relevant to the questioner's goal.

To implement expected relevance of an answer, the QA model builds on decision-theoretic accounts of relevance of questions and answers \citep{van2003questioning, benz2006utility},  
which formalizes relevance in terms of a \textit{decision problem (DP).}
The DP includes a real-valued \textit{utility function} of how useful different alternatives (e.g., iced coffee, soda, Chardonnay) are for a given goal (e.g., getting an iced tea).
The questioner selects questions that have a high expected relevance (i.e., high \textit{expected utility}) of information from the base-level respondent.
The pragmatic respondent uses the questioner’s goal-oriented choice of question to infer from the question what kind of DP the questioner likely has. 
These inferences then guide the respondent's choice of information that will likely increase the expected utility for the questioner, traded off with response costs. 
We use a probabilistic implementation of the QA model in WebPPL \citep{dippl} from \citet{hawkins2025relevant} as a starting point and baseline.
As commonly done for probabilistic modeling, for these simulations we specified the space of possible answers, possible questions, the literal semantics and the DP utility function specifically for the main experimental materials  (see Section~\ref{sec:procedure} and Appendix~\ref{sec:app:qa-model-params}). 

Before diving into neuro-symbolic model evaluation, we first validate whether the task decomposition stipulated in the QA model is actually borne out in human intuitive reasoning.
To this end, we conducted an exploratory \textit{answer explanation} experiment.
Participants (N=50) were recruited via Prolific and shown four trials with contexts wherein a person asked for a target item while several alternative options were available, similar to the initial café example, which constituted the main materials we describe in more detail in Section~\ref{sec:procedure}. 
The question was followed by a character replying ``no'' and providing one, most relevant, competitor alternative. 
Participants were asked to type an explanation of why that response was reasonable and what would justify mentioning the particular option over a different one.
We then analyzed the types of provided explanations, distinguishing between explanations that appealed to (1) abstract similarity of options, (2) questioner goals, desires, intentions, or preferences, and (3) features that were functionally relevant for the questioner goal (e.g., being and iced non-alcoholic drink). 
If participants spontaneously reason about questioner goals and respective relevant option features as formalized in the QA model, we hypothesize that the proportion of (2) and (3) will be higher than (1).
We found that 0.43 of responses appealed to goals (2), 0.20 to goal-relevant features (3), and 0.21 to general similarity (1). 
0.13 of responses were unclassifiable (e.g., only appealed to respondent politeness).
We interpret this as mild \textit{prima facie} support for the task decomposition implemented in the probabilistic QA model.
In the next section, we analyze how systematically replacing different components of the QA model  with LLM modules affects the fit to human data.

\begin{figure*}[t]
	\centering
	\includegraphics[width=0.95\textwidth]{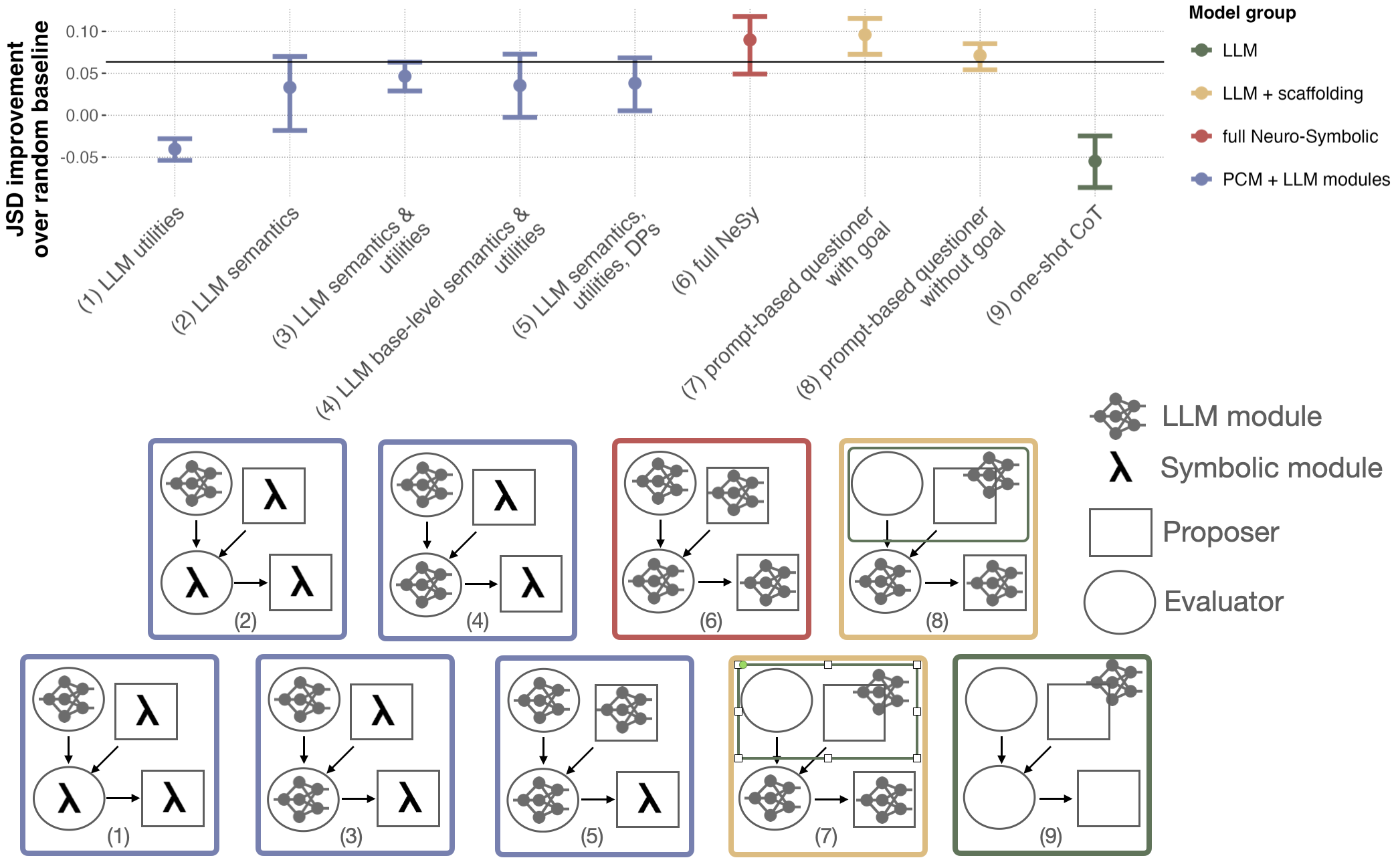}
	\caption{\label{fig:models-jsd}\textbf{ Upper panel:} Improvement of the model fit to human data in terms of Jensen-Shannon divergence over a uniform response distribution baseline (higher is better, $y$-axis) of all analyzed models ($x$-axis). 
    The horizontal line indicates performance of the probabilistic model. 
    Dots indicate the means across simulations, error bars indicate 95\% bootstrapped CIs. 
    \textbf{Lower panel:} Overview of tested models. 
    Each box shows a schematic of one model, labeled on the $x$-axis in the plot above it. 
    The models are ordered from closest to the PCM on the left (only one component is LLM-based), to a model only using a single LLM with a single prompt on the right. 
    }
\end{figure*}

\section{Evaluating Neuro-Symbolic QA models}
\label{sec:open-ending}

We investigate the neuro-symbolic framework starting with models where only one component of the task is supplied by an LLM.
We then incrementally increase the number of LLM-based modules and change their types, while observing the changes of the \textit{fit to human data} and the \textit{qualitative changes in the predictions}.
The driving motivation is to make PCMs more generally applicable (open-ended).
For that, two steps are necessary.
For one, we would like to be able to generate an in principle open-ended set of alternatives over which to reason or which to choose from. 
Consequently, we test if LLMs can provide plausible sets of responses, questions, and questioner goals for the QA model; we call LLMs in this role \textbf{proposers} \citep[cf.][]{sumers2023cognitive, tsvilodub2024cognitive}. 
For another, once we have open-ended sets of alternatives, we need to be able to obtain information about them for down-stream computation, i.e., we also use LLMs in the role of \textbf{evaluators} for judging literal semantics of answers and for assessing the utility of options.

\subsection{Experimental setup} 
\label{sec:procedure}
For all reported simulations below, we use \texttt{GPT-4o-mini} for the LLM modules, with the sampling temperature $\tau=0.1$. All simulations are run for five iterations. We report additional results with the open-source LLM \texttt{Qwen-2.5-32B-Instruct} in Appendix~\ref{sec:app:qwen-results}.
We use experimental materials, human data and the one-shot LLM prompt from \citet{tsvilodub2023overinformative} to investigate what kinds of alternative options (e.g., iced coffee or Chardonnay), if any, different neuro-symbolic QA models mention in the predicted responses, given a polar question (e.g., “Do you have iced tea?”) and different options in context. 

The materials include 30 commonsense vignettes similar to the initial barista example. 
The context always included three possible options, but not the requested target (i.e., iced tea). 
The options always included a best-fitting alternative called the \textit{competitor} (e.g., iced coffee), a conceptually \textit{similar} option that was deemed less relevant for the questioner's goal (e.g., soda), and an \textit{unrelated} option irrelevant for the uttered request (e.g., Chardonnay). 
Experimental subjects provided answers by freely typing into a text box.
Responses were categorized as ``target,'' ``similar,'' and ``unrelated.''
In addition to these three categories, corresponding to mentioning each of the single options, the categorization also distinguished responses that mentioned \textit{all options}, as well as responses that mentioned \textit{no options}. 

If a respondent is engaging in pragmatic reasoning, we would expect her to prefer competitor responses over other types.
\citet{tsvilodub2023overinformative} found that humans are, in fact, \textit{relevantly overinformative}, strongly preferring competitor responses (0.52 of responses) over exhaustive responses (0.10), no options responses (0.20), similar (0.18) or unrelated responses (0.00). 
We investigate how well neuro-symbolic models match human behavior, operationalized via Jensen-Shannon divergence between the observed human data and the models' categorical predictions.

\subsection{Integrating LLM Evaluators in the PCM}
\label{sec:pcm-llm-evaluators}
We assess a class of models that, starting from the QA model, systematically incorporate LLM modules into the PCM architecture which take over two functions: (i) the evaluation of utility of an option, and (ii) the evaluation of the truth of a response. 
Figure~\ref{fig:models-jsd} (lower panel) shows a schematic overview of the tested models. 

First, we implement an LLM \textit{utility evaluator} for instantiating the utility function in the questioner's decision problem (resulting in the \textbf{``LLM utilities'' model}). 
The utility function defines real-valued utilities for the different alternatives (e.g., the iced coffee, soda), conditioned on a target object (e.g., iced tea).
In the original QA model, the utilities were elicited in a human rating experiment wherein participants were asked to provide slider ratings for each possible option (e.g., iced tea, iced coffee, soda, Chardonnay), given another option as the goal (see Appendix~\ref{sec:app:qa-model-params}). 
To replace the human input with an LLM, we prompted the utility evaluator in a way identical to the instructions of the human elicitation experiment, namely to predict the full space of utilities via ratings on a scale with range 0--100 instead of slider ratings. 
Importantly, the prompt (and the original human experiment) only asked for abstract ratings, independent of the functional context in which the options occurred in the question answering scenario
(see Appendix~\ref{sec:app:prompts} for all full prompts). The remaining model components (e.g., the set of alternative utterances, the semantics) remained symbolic in this model. 

Beyond replacing the utility component, another function-based component to replace with LLMs for open-ending the PCM is \textit{semantic evaluation}.
Semantic evaluation is necessary for the base-level and for the pragmatic respondent and assesses whether a response is true in a particular context.
While base-level and pragmatic respondent have slightly different responses at their disposition owing to the fact that the base-level responder is not reasoning about the context (see Appendix~\ref{sec:app:qa-model}), the semantic evaluation is essentially the same.
For an answer like ``No, but we have iced coffee.'' the module has to check whether the polar answer part (e.g., ``yes'', ``no'') is true for a context (e.g., the caf\'e has soda and iced coffee), given the question (e.g., ``Do you have iced tea?'').
It also has to evaluate whether the added information (e.g., ``We have iced coffee.'') is actually correct.
We explored models with different combinations of these evaluators. 
The \textbf{``LLM semantics''} model uses an LLM-based semantic evaluator for both the base-level and the pragmatic respondent, while using the same utility component as the original QA model (based on the human experimental data).
The \textbf{``LLM semantics \& utilities''} model employs all described LLM evaluators. The \textbf{``LLM base-level semantics \& utilities''} only uses an LLM-based base-level respondent, a rule-based pragmatic respondent, and the LLM utility evaluator.  
The predictions of all models are compared in Section~\ref{sec:results}.

\subsection{Integrating LLM Proposers in the PCM}
\label{sec:pcm-llm-proposers}
Next, we integrate LLMs as \textit{proposers} for sets of alternatives required by the QA model. 
We start with sampling the possible questioner goals with a \textit{goal proposer}. 
The LLM was prompted to generate plausible text-based goals, given the context and question (see~Figure~\ref{sec:app:prompts-goal-proposer}). 
While the set of possible goals in the PCM only contained four DPs (each defining a preference for one of the options: target, competitor, similar, unrelated option), the proposer may sample any text-based questioner goal description. 
These sampled text-based goals are connected to a DP representation via the \textit{utility evaluator} (Section~\ref{sec:pcm-llm-evaluators}). The evaluator was prompted to generate the utilities for the available options, conditioned on each proposed goal.
The \textbf{``LLM semantics, utilities, DPs''} model uses the goal proposer together with the evaluators from Section~\ref{sec:pcm-llm-evaluators}, while the sets of possible utterances and questions are symbolic (i.e., pre-specified manually).  

Further open-ending the QA model, we introduce a \textit{response proposer} and a \textit{question proposer} which provide the set of alternative questions and pragmatic answers that the respective pragmatic agents reason over. 
In both cases, the LLM was concisely prompted to generate $n$ alternatives to an observed utterance or question given the context vignette (see~Figure~\ref{sec:app:prompts-response-proposer}, Figure~\ref{sec:app:prompts-question-proposer}). 
We set $n = 10$ for the response proposer, and $n=3$ for the question proposer.
Here, we address the empirical question whether LLMs, out of the box, can be (easily) prompted to produce the expected types of alternative pragmatic responses in the context of the QA model (no options, competitor, similar, unrelated, all options). Based on exploratory qualitative analyses described in Section~\ref{sec:results} in more detail, we append ``no-options'' and ``all-options'' responses constructed in a rule-based manner to the set of sampled alternatives.
The observed question was always added to the set of sampled alternatives provided by question proposer.

The question and response proposers were tested as part of the fully neuro-symbolic replication of the PCM (\textbf{``full NeSy''} model).
This model implements the full task decomposition of the QA model, capturing the pragmatic respondent's recursive reasoning (Figure~\ref{fig:prior-pq}) fully via the modules described above. 
The base-level respondent uses an LLM-based semantic evaluator to (symbolically) select an informative, true response to a given question (assuming that the decision problem is known).  
For the pragmatic interpreter, the different possible questions are supplied by an LLM-based question proposer.
An LLM-based utility evaluator rates the usefulness of potential options to (symbolically) compute the questioner's expected utility of each question (based on the expected behavior of the base-level respondent). 
Finally, the pragmatic respondent estimates likely DPs among the neurally sampled alternatives, given the question, symbolically via Bayes rule (where the likelihood term is approximated via samples of generated questions given a DP).
Given her posterior beliefs about the DPs, the respondent chooses a response from the set provided by the response proposer that maximizes her utility function. 
The respondent's utility function combines the expected utility of a response with informativeness, formalized as a KL divergence term (see Appendix~\ref{sec:app:qa-model} for details).
We assume flat priors and no utterance costs throughout the model.

\subsection{Scaffolding Prompted LLMs with Cognitive Modules}
\label{sec:llm-scaffolding}
All previous models have implemented computational components suggested by the original QA model with LLM-based proposers and evaluators.
These LLM-based components implemented rather ``local'', smaller computational elements of the task analysis suggested by the QA model. 
Alternatively, we may also use LLMs to replace larger chunks of computation, such as the full pragmatic question answering agent, or even the full task analysis captured by the QA model.
In the following, we introduce three models that instantiate this general strategy.


We first consider a model called \textbf{prompt-based questioner}, of which we consider two versions, one prompted with questioner goals, and one prompted without goals. 
This model decomposes the pragmatic respondent's task into its two high-level components suggested by the PCM: inferring the questioner's goal based on the observed question, and selecting a response that optimizes the questioner's utility given the inferred DP.
We implement a purely prompt-based pragmatic questioner module that supplies the first component. 
This prompt-based questioner is used by the pragmatic respondent of the ``full NeSy'' model  for inferring the distribution over DPs sampled with an LLM-based goal proposer. 
The prompt-based questioner takes a questioner goal, the context, and prompts the LLM to provide a likelihood of someone asking the given question (see Fig.~\ref{sec:app:prompts-folded-q}). The elicited likelihoods for all questions and DPs are then renormalized and used by the pragmatic respondent.
We then compare the role of conditioning this module on the goal, and also use a goal-free prompt where the LLM is asked to assess the question likelihood based on the context only (\textbf{prompt-based questioner without goal}, see Fig.~\ref{sec:app:prompts-folded-q-noGoal}).

For comparison, we also consider a purely \textit{monolithic} prompting of the LLM.
In particular, the \textbf{one-shot chain-of-thought model} has a chain-of-thought prompt which \textit{verbalizes} the reasoning steps suggested by the QA model in the chain-of-thought for a single example item (see Figure~\ref{sec:app:prompts-cot}). That is, this model is fully LLM-based, using only one call to one neural module (i.e., the LLM).

\section{Results}
\label{sec:results}

\begin{figure*}[t]
	\centering
	\includegraphics[width=0.95\textwidth]{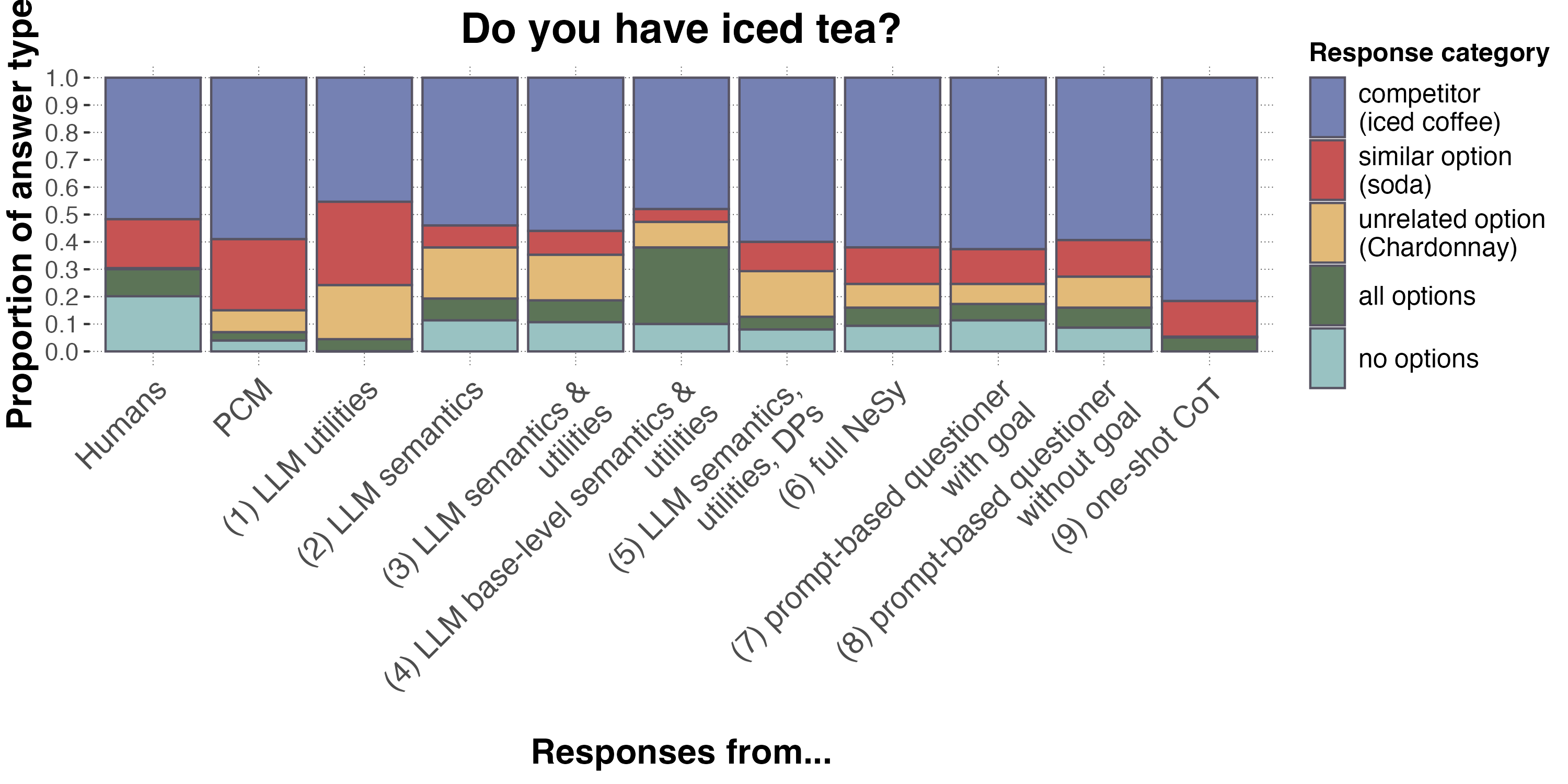}
	\caption{\label{fig:response-props} Proportions of different response categories produced by humans (left column) and predicted by different models. The categories are based on which options are mentioned in the response.}
\end{figure*}

\paragraph{Quantitative results} We used the human answer proportions reported in Section~\ref{sec:open-ending} as reference and quantitatively compared models in terms of fit to the human data by calculating the Jensen-Shannon divergence ($\mathit{JSD}$) between the human and the models' predictions. 
Specifically, we calculated the score $\Delta_i$ of model $M_i$ in comparison to the performance of a baseline $B$ given by a flat distribution over all answer categories: 
$$
\Delta_i =  \mathit{JSD}(B, \text{humans}) - \mathit{JSD}(M_i, \text{humans})
$$
where $\mathit{JSD}(B, \text{humans}) = 0.154$. 
We report $\Delta_i$-s in Figure~\ref{fig:models-jsd} (upper panel; higher JSD differences are better, indicating closer fit to human data). 
The figure additionally shows the reference value provided by the PCM (solid line).  

We found that most tested models with intermediate or high degrees of task decomposition came close to the original PCM (the CIs overlap with the PCM reference line or lie above it), indicating that the neuro-symbolic framework provides a potentially viable method for explaining human data.
Visually, the ``full NeSy'' model and the ``prompt-based questioner with goals'' fit human data best in terms of $\Delta$.
The PCM + LLM models tended to improve with a higher number of LLM modules, but generally provided a somewhat worse fit than the PCM (the means are below the line). 
Supporting LLMs with a theoretically motivated task decomposition led to significant improvement within the LLM + scaffolding models: the ``prompt-based questioner'' models showed a better fit than the ``one-shot CoT'' model. 
Therefore, overall we found that the neuro-symbolic approach to open-ending pragmatic PCMs showed quantitative fit to human data on par with established cognitive modeling, while offering a more realistic interface to natural language inputs and outputs.

\paragraph{Qualitative results} Next to the quantitative analyses, we analyzed qualitatively the differences between model predictions and the performance of the single modules.
Figure~\ref{fig:response-props} shows the proportions of different response categories (e.g., competitor, no-options responses etc.) predicted by the different models, next to PCM predictions and human data from \citet{tsvilodub2023overinformative}.
The figure reveals that although many neuro-symbolic models have similar fit to human data in terms of $\Delta$, there are qualitative differences in the predicted response proportions. 
The two models with ``LLM semantics'' overpredicted the proportion of unrelated responses, while the ``LLM base-level semantics \& utilities'' model overpredicted the all-options response rate and slightly underpredicted the competitor rate. 

Comparisons of the base-level and pragmatic respondent semantic modules revealed that the base-level semantics module performed reliably, while the pragmatic respondent semantic module made mistakes more frequently, including when evaluating unrelated responses. This may have led to the overprediction of the unrelated responses, as shown by the comparison of the ``LLM semantics \& utilities'' and the ``LLM base-level semantics \& utilities'' models because the former only differs from the latter by using an LLM-based pragmatic respondent semantics evaluator. 
We correlated the utility evaluator predictions with data elicited from humans for the PCM (see Figure~\ref{fig:llm-utils}) and found a very high correlation ($R=0.92$), so we can likely rule out the utility evaluator as the source of overprediction of the unrelated category.

The comparison of the PCM + LLM models to the ``full NeSy'' model highlights the difference in response proportions that is driven by adding LLM proposers for the set of available responses and questions. The addition of response and question proposers decreased the rate of unrelated responses and slightly increased the rate of similar and exhaustive responses. 
Since the ``full NeSy'' model included the pragmatic respondent semantic evaluator module, we can conclude that semantic evaluations might work more reliably with the LLM's own proposals than with the pre-specified sets of responses and questions. 
These observations are in line with one of the well-known challenges of neuro-symbolic modeling concerning difficulty of converting between neural and symbolic representations that is required in order to reliably compute truth values for open-ended sentences and contexts \citep{bader2004integrationconnectionismfirstorderknowledge}, as well as with debates around LLMs' ability to provide reliable evaluations \citep[][]{bavaresco2406llms}.

We also explored decreasing and increasing the $n$ of alternative responses proposed by the LLM. 
We found that results with $n<10$ proposals were unlikely to contain the ``all options'' or ``no options'' responses.
For $n=10$ this was more often the case, but we appended these two response types to set of alternatives manually nonetheless, to ensure availability of all conceptually meaningful response types. 
Sampling $n=50$ responses ensured full coverage of response types but became computationally expensive. 
Generally the proposals often contained multiple instances of one response type (e.g., multiple competitor responses), an observation we return to in the discussion.
However, this is unlikely the sole driving force beyond the fit of the framework, as the ``LLM semantics, utilities, DP'' model showed a similar competitor response proportion, while operating on a fully prespecified set of responses. 

We qualitatively assessed the samples of the goal proposer module that generates possible text-based questioner goals, given the vignette. 
We compared the samples to human data from a web-based experiment wherein participants were asked to write three plausible goals of the questioner, given the vignette context (see~Appendix~\ref{sec:app:goal-sampling-expt} for details and human results). We focused on analyzing whether the LLM-proposed goal focused on getting the \textit{target} mentioned in the question, on a more \textit{general} information gain, or on \textit{specific} situation aspects.
We observed that, while LLM proposals were plausible, they focused on the target and specific goals around the target more, while humans showed more diversity in their specific goals, e.g., often involving social aspects of the described situation.

Turning to the LLM + scaffolding model type, comparing the ``prompt-based questioner model without goals'' and the ``prompt-based questioner model with goals'' revealed a trend towards predicting unrelated and similar responses more uniformly in the goal-free model, which is expected given that the distinction between these types of answers is based on reasoning about the questioner's goal. 
However, these differences are small and indicate that, even under certain (ablating, from a theoretical perspective) prompt variation, LLMs may be able to approximate pragmatic behavior.

Taken together, our key results are:
\begin{itemize}
    \item the neuro-symbolic modeling approach fits human data quite closely, potentially making it a framework for computational modeling of pragmatic question answering performing on par with the PCM;
    \item at least some level of task decomposition when using LLM modules is required for a good fit to human data;
    \item LLM modules are generally good proposers, although attention should be paid to \textit{types} of proposals that are expected for explanatory purposes;
    \item LLMs are good evaluators for functions based on abstract world knowledge like the utility evaluator;
    \item LLMs may struggle with truth-conditional semantics of certain utterances, but perform well when evaluating yes/no responses to polar questions. 
\end{itemize}



\section{Related work}
\label{sec:related-work}
Our work is situated at the intersection of several strands of like-minded work in different areas, in addition to the work we build on directly \citep{Hawkins2015WhyDY, tsvilodub2023overinformative}.
The idea and promise of neuro-symbolic models has been studied in artificial intelligence for many years \citep{bhuyan2024neuro}. 
Further, our framework is closely related to recent work outlining various approaches to combining scaffolding structures, computational modeling or cognitive architectures with LLMs \citep[e.g.,][]{nye2021improving, collins2022structuredflexiblerobustbenchmarking, sumers2023cognitive, Wong2023FromWM, kambhampati2024llmscantplanhelp}.
Combining LLMs with PCMs specifically in the context of computational pragmatics has received some attention in recent work \citep[e.g.,][]{lew2020leveraging, franke2024bayesian, tsvilodub2024cognitive} but the present work focuses specifically on systematically comparing and evaluating families of related models with varying degrees of neural or symbolic computation. 

On an algorithmic level, our models combine several LLM calls in a particular architecture, which has been widely used in recent prompt techniques \citep{nye2021improving, prystawski2023psychologically, yao2023treethoughtsdeliberateproblem}, and systems that use LLM calls to retrieve information \citep[e.g.,][]{lewis2020retrieval}, to access different tools \citep[e.g.,][]{schick2023toolformer} or to solve complex reasoning tasks \citep[e.g.,][]{creswell2022selection, he2023solving}.

Systems with multiple LLM calls per input have also been specifically applied to question answering \citep{wang2023selfconsistencyimproveschainthought}, mainly with a focus on improving factual accuracy of responses, or on training systems to improve their question asking capabilities \citep{andukuri2024stargate}.
Therefore, our case study addresses a highly relevant task, with a novel focus on modeling \textit{pragmatic, human-like} answering behavior. 

\section{Discussion}
\label{sec:discussion}
Taken together, in this case study we outlined and systematically assessed a neuro-symbolic framework for computational pragmatic modeling that uses probabilistic cognitive models as scaffolding structure that integrates LLM components for more flexible interfaces with language and background knowledge. The experiments on a case study of pragmatic question answering revealed that such modeling can be a viable candidate in the toolbox for more flexible models of human behavior in question answering.
The systematic comparison of neuro-symbolic models with different degrees of task decomposition suggests fine-grained differences in how LLMs perform on different subtasks common to PCMs. 

Our case study has several limitations, but also opens up paths for future work. 
For one, the full neuro-symbolic models implement Bayesian inference via enumeration, which results in computational bottlenecks when scaling the number of proposals and options in context.
Related work connecting LLMs and Bayesian inference might be a promising avenue for improvements \cite{lew2023sequentialmontecarlosteering}.
Additionally, the current main results are based only on one closed-source LLM (but see Appendix~\ref{sec:app:qwen-results} for exploratory results with an open-source LLM), and only use zero-shot prompting (except the CoT model). 
In this initial case study, we prioritized using relatively simple, non-engineered prompts, but nonetheless LLM prompting comes with potential risks of hallucination, errors and biases \citep[e.g.,][]{bender2021stochastic, ji2023hallucination, liu2023lost}. 

Finally, the use of LLMs as proposers and evaluators opens up interesting questions. For instance, response proposals supplied by the LLM might contain a trend towards certain response types, which can arguably be seen as a learned prior over human preferences reflected in the training data. 
Additionally, cognitive models usually assume utterance costs for human language production and comprehension, but such online processing costs might not have a clear counterpart in LLMs. Further, varying performance of LLM evaluators might suggest that some aspects of semantics might be amortized in training data \citep{white2020learningreferinformativelyamortizing}.
Our results suggest that LLMs might not approximate different aspects of human intuitive knowledge equally well, touching upon important considerations of replacing human judgements with LLMs \citep{ShiffrinMitchell2023:Probing-the-psy, lohn-etal-2024-machine-psychology}.
For the LLMs + PCM models, one other potential source of improved performance with scaffolding of the LLM could be due to higher inference time compute budget that comes with decomposing the task into several LLM calls \citep{yu2024distilling21}. 

In sum, we presented a detailed case study as a starting point for exploring neuro-symbolic models of human language use, showing that task decomposition supplied by a cognitive model can be leveraged in synergy with recent LLMs, working towards open-ending pragmatic computational modeling.

\section*{Acknowledgments}
We would like to thank Fausto Carcassi for his contributions to developing the framework, and the anonymous reviewers for insightful comments.
MF is a member of the Machine Learning Cluster of Excellence at University of T\"ubingen, EXC number 2064/1 – Project number 39072764.
PT and MF gratefully acknowledge the support by the state of Baden-W\"urttemberg through bwHPC and the German Research Foundation (DFG) through grant INST 35/1597-1 FUGG.
\bibliography{references}

\appendix
\label{sec:appendix}

\begin{figure*}[t]
	\centering
	\includegraphics[width=0.75\textwidth]{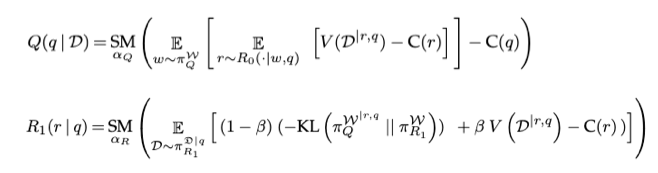}
	\caption{\label{fig:ppq-math} Formal definitions of the pragmatic questioner $Q(q \mid D)$ and respondent $R_1(r \mid q)$.}
\end{figure*}
\section{QA model}
\label{sec:app:qa-model}
Below, we report the QA model by \citet{hawkins2025relevant}, described in Section~\ref{sec:prior-pq}, in more formal detail.

The base-level respondent that provides literal responses $r$ to a question $q$ given the world $w$ is defined as follows:
\begin{align*}
    R_{0}(r \mid w, q)  \propto 
      \begin{cases}
        1 & \text{if $r$ is true in $w$ \& safe for $q$} \\
        0 & \text{otherwise.}
  \end{cases}
\end{align*}
The notion of safety is couched in prior work on semantics of questions and answers \citep{PruittRoelofsen2011:Disjuntive-ques} and entails that, for the tested vignettes, only the literal answers $r\in$ \{`yes', `no'\} are evaluated here. 

The pragmatic questioner selects a question given their decision problem, based on the responses they expect from the base-level respondent $R_0$.
Formally, a \textit{decision problem} (DP) is a tuple $\mathcal{D} = \tuple{\mathcal{W}, \mathcal{A}, \mathcal{U}, \pi_{Q}^{\mathcal{W}}}$,  consisting of
a set of world states $\mathcal{W}$,
a set of options $\mathcal{A}$,
a utility function $\mathcal{U} \mycolon \mathcal{W} \times \mathcal{A} \rightarrow \mathds{R}$, and 
a probability distribution $\pi_{Q}^{\mathcal{W}} \in \Delta(\mathcal{W})$ capturing the questioner's prior beliefs about the world states.
%
Then, the \textit{value of a decision problem} $\mathcal{D}$ is the expected utility under a policy $\aleph^{\mathcal{D}}$ that chooses options according to their expected utility:
\begin{align*}
  V(\mathcal{D}) = \mathop{\mathbb{E}}_{a \sim \aleph^{\mathcal{D}}} \left[\, \mathop{\mathbb{E}}_{w \sim \pi_{Q}^{\mathcal{W}}} \Big[\,\mathcal{U}(w,a) \,\Big] \, \right]
 \end{align*}

The pragmatic questioner then selects a question by soft-maximizing the expectation over the values of the decision problems $\mathcal{D}^{\mid  r,q}$ given likely responses from the base-level respondent, resulting in $Q(q \mid D)$ (see Figure~\ref{fig:ppq-math}),
where \text{C}(r) and \text{C}(q) are the production costs associated with the response and question, respectively. 

The pragmatic respondent then reasons about the pragmatic questioner's choice of question in order to infer their likely decision problem: 
\begin{align*}
  \pi^{\mathfrak{D} \mid q}_{R_{1}}(\mathcal{D}) 
  \propto
  Q(q \mid \mathcal{D}) \ \pi^{\mathfrak{D}}_{R_{1}}(\mathcal{D}) 
\end{align*}

Finally, the pragmatic respondent chooses a response by soft-maximizing the expected utility of the response given their posterior beliefs about the questioner DP.
Utility is defined as a (parameterized) combination of informativity (defined via KL divergence) and action-relevance (defined via the decision problem value), resulting in $R_1(r \mid q)$ (see Figure~\ref{fig:ppq-math}).

\subsection{Parameterization of the QA model}
\label{sec:app:qa-model-params}
As commonly done for probabilistic modeling, in order to run simulations with the QA model parameters of the model were specified by the modelers or with elicited human data \citep{hawkins2025relevant}.
For each vignette, the set of alternative questions included polar questions about the availability of each of the possible options individually, and a wh-question inquiring about all possible options. 
The set of available pragmatic answers included answers of all categories described in Section~\ref{sec:procedure}.

\begin{figure*}[t]
	\centering
	\includegraphics[width=0.95\textwidth]{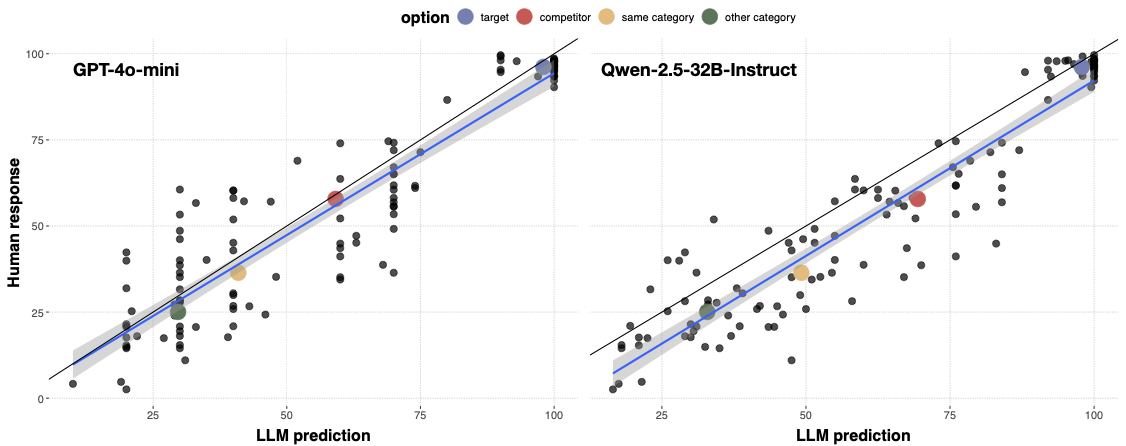}
	\caption{\label{fig:llm-utils} \textbf{Left}: GPT-4o-mini utilities plotted against human utilities, $R=0.92$. \textbf{Right}: Qwen-2.5-32B-Instruct utilities plotted against human utilities, $R=0.93$.}
\end{figure*}

In order to specify the utility functions of the questioner DPs, a web-based experiment was run with human participants. 
Participants ($N=453$) were asked to provide slider ratings for each possible option (e.g., iced tea, iced coffee, soda, Chardonnay), given another option as the goal. The full space of possible combinations was elicited. The slider ratings were on a scale of 0--100. Importantly, participants were asked to rate how happy they think a person would be to receive an option, given the target, resulting in \textit{abstract} conditional preferences.
The DP utilities for each vignette were bootstrapped from human preferences in the QA model simulations.
Human results for ratings of the alternatives, given the option used as the target in the free production experiments as the goal (e.g., the iced tea) are shown in Figure~\ref{fig:llm-utils} (left) together with respective LLM module predictions. Human and \texttt{GPT-4o-mini} ratings correlated highly, and supported the intuitive ordering of the relevance of alternatives (e.g., the competitor received higher ratings than the unrelated option for a given target).

\section{Prompts}
\label{sec:app:prompts}
Prompts for all LLM modules are presented below in Figures~\ref{sec:app:prompts-utility-eval}--\ref{sec:app:prompts-cot}.
\begin{figure}[htpb]
\centering
\begin{tcolorbox}[
width=1\linewidth,
title={Utility Evaluator Prompt}]
\fontsize{5pt}{5pt}\selectfont
\ttfamily
\begin{lstlisting}[language={}]
In this study we are interested in how you think about other people.
On each trial, you will be given some information about a person: 'Suppose someone wants to have Italian food.'

Then we'll ask how happy you think this person would be about other things, given this information. For instance, we might ask: 'How happy do you think they would be if they had French food instead?'
You'll use ratings from 0-100 to answer the questions. Return the rating only.

Suppose someone wants {goal}. How happy do you think they would be if they got {option}?
\end{lstlisting}
\end{tcolorbox}
\caption{\textbf{Utility Evaluator Prompt}
\label{sec:app:prompts-utility-eval}}
\end{figure}

\subsection{Semantic Evaluators}
\label{sec:app:prompts-semantic-eval}
\begin{figure}[htpb]
\centering
\begin{tcolorbox}[
width=1\linewidth,
title={Base-level Evaluator Prompt}]
\fontsize{5pt}{5pt}\selectfont
\ttfamily
\begin{lstlisting}[language={}]
Safe answers to questions only provide information that the questioner genuinely does not know, given what they asked.
True answers to questions only provide information that is true given the context.

Here is an everyday situation where someone asks a question: {context + question}
Here is a potential answer to the question: {utterance}

Is the answer safe and true in this context, according to the definition above?
Return 'yes' or 'no' only.
\end{lstlisting}
\end{tcolorbox}
\caption{\textbf{Base-level Evaluator Prompt}
\label{sec:app:prompts-semantic-eval-base-level}}
\end{figure}
\begin{figure}[htpb]
\centering
\begin{tcolorbox}[
width=1\linewidth,
title={Pragmatic Respondent Semantic Evaluator Prompt}]
\fontsize{5pt}{5pt}\selectfont
\ttfamily
\begin{lstlisting}[language={}]
True answers to questions only provide information that is true given the context.

Here is an everyday situation where someone asks a question: {state}
Here is a potential answer to the question: {utterance}

Is the answer true in this context, according to the definition above?
Return 'yes' or 'no' only.
\end{lstlisting}
\end{tcolorbox}
\caption{\textbf{Pragmatic Respondent Semantic Evaluator Prompt}
\label{sec:app:prompts-semantic-eval-pragmatic}}
\end{figure}
The base-level semantic evaluator only evaluates the set of literal responses \{`yes', `no'\}. The pragmatic respondent semantic evaluator evaluates the set of possible overinformative responses. In models where the set of pragmatic responses is pre-specified, the possible responses are of the form  ``I'm sorry, we don't have \{target\}. \{continuation\}'', where the continuation was constructed for all response types (no-options, competitor, similar, unrelated, all-options responses).

\begin{figure}[htpb]
\centering
\begin{tcolorbox}[
width=1\linewidth,
title={Response Proposer Prompt}]
\fontsize{5pt}{5pt}\selectfont
\ttfamily
\begin{lstlisting}[language={}]
Safe answers to questions only provide information that the questioner genuinely does not know, given what they asked.
True answers to questions only provide information that is true given the context.

Here is a question someone could ask in an every day situation: {question}
Here are the available options: {options}

Generate {num_samples} literal answers to the question.
Return them as a numbered list.
\end{lstlisting}
\end{tcolorbox}
\caption{\textbf{Response Proposer Prompt}
\label{sec:app:prompts-response-proposer}}
\end{figure}

\begin{figure}[htpb]
\centering
\begin{tcolorbox}[
width=1\linewidth,
title={Question Proposer Prompt}]
\fontsize{5pt}{5pt}\selectfont
\ttfamily
\begin{lstlisting}[language={}]
Suppose a person has the following goal: {goal}
The person is in the following everyday situation: {context}
Generate {num_samples} well formed short questions(s) the person might naturally ask in the context to achieve their goal.
                          
\end{lstlisting}
\end{tcolorbox}
\caption{\textbf{Question Proposer Prompt}
\label{sec:app:prompts-question-proposer}}
\end{figure}

\begin{figure}[htpb]
\centering
\begin{tcolorbox}[
width=1\linewidth,
title={Goal Proposer Prompt}]
\fontsize{5pt}{5pt}\selectfont
\ttfamily
\begin{lstlisting}[language={}]
You will be given a context in which a person asks a question.
What plausible different goals might the person be interested in, given what they asked?
Your task is to generate {num_samples} alternatives in a comma separated list.
                          
\end{lstlisting}
\end{tcolorbox}
\caption{\textbf{Goal Proposer Prompt}
\label{sec:app:prompts-goal-proposer}}
\end{figure}

\begin{figure}[htpb]
\centering
\begin{tcolorbox}[
width=1\linewidth,
title={Prompt-based questioner with goals}]
\fontsize{5pt}{5pt}\selectfont
\ttfamily
\begin{lstlisting}[language={}]
We are interested in how likely a person would be to ask the following question in a simple context, given their goal. 
Please return only the likelihood, provided on a scale between 0 and 1.
Goal: {goal}
Context: {state}
{utterance}
\end{lstlisting}
\end{tcolorbox}
\caption{\textbf{Prompt-based questioner with goals}
\label{sec:app:prompts-folded-q}}
\end{figure}
\begin{figure}[htpb]
\centering
\begin{tcolorbox}[
width=1\linewidth,
title={Prompt-based questioner without goals}]
\fontsize{5pt}{5pt}\selectfont
\ttfamily
\begin{lstlisting}[language={}]
We are interested in how likely a person would be to ask the following question in a simple context. 
Please return only the likelihood, provided on a scale between 0 and 1.
Context: {state}
{utterance}
\end{lstlisting}
\end{tcolorbox}
\caption{\textbf{Prompt-based questioner without goals}
\label{sec:app:prompts-folded-q-noGoal}}
\end{figure}
\begin{figure}[htpb]
\centering
\begin{tcolorbox}[
width=1\linewidth,
title={One-shot chain-of-thought prompt}]
\fontsize{5pt}{5pt}\selectfont
\ttfamily
\begin{lstlisting}[language={}]
You are hosting a barbecue party. You are standing behind the barbecue. You have the following goods to offer: pork sausages, vegan burgers, grilled potatoes and beef burgers. 
Someone asks: Do you have grilled zucchini? 

Let's think step by step. You reason about what that person most likely wanted to have. That they asked for grilled zucchini suggests that they might want vegetarian food. From the items you have pork sausages and beef burgers are least likely to satisfy the persons desires. Vegan burgers and grilled potatoes come much closer. Grilled potatoes are most similar to grilled zucchini. You reply: 

I'm sorry, I don't have any grilled zucchini. But I do have some grilled potatoes.
                          
\end{lstlisting}
\end{tcolorbox}
\caption{\textbf{One-shot chain-of-thought prompt}
\label{sec:app:prompts-cot}}
\end{figure}

\section{Human Experiment on Goal Inference}
\label{sec:app:goal-sampling-expt}
In an exploratory \textit{goal inference} study, participants (N=35) were shown vignette contexts without the available options, followed by the question asked by a speaker.
Participants were asked to name three plausible goals in three separate text fields that the questioner might have in mind when asking the question. 
We focused on distinguishing whether participants named goals focused on acquiring the \textit{target} mentioned in the question, on acquiring more \textit{general} information, or on goals related to more \textit{specific} aspects of the situation. 

Participants were most likely to infer \textit{specific} goals (0.42 of the responses), followed by \textit{target-related} goals (0.35 of the responses). More \textit{general} information-seeking goals were less likely (0.17 of the responses), and some responses were non-classifiable (0.06).

We then manually analyzed the proposals of the LLM goal proposer module.
Qualitatively, the target-related goals mostly were about acquiring the target or an item with the same functional features (e.g., when the target was veggie pizza, the functional feature would be being a vegetarian option), both for humans and LLMs.  
The specific goals produced by humans often involved more details than just acquiring the target, e.g., acquiring the target for a friend, or mentioned different specific preferences participants came up with. In contrast, the specific goals produced by LLMs were less likely to mention social aspects like acquiring something for a friend, and more likely to produce possible more specific questioner preferences (e.g., ``asking about certain dietary restrictions'').
The more general goals produced by humans and LLMs often mentioned learning about the set of available alternatives.

\section{Simulation Results with an Open-Source LLM}
\label{sec:app:qwen-results}
\begin{figure*}[t]
	\centering
	\includegraphics[width=0.95\textwidth]{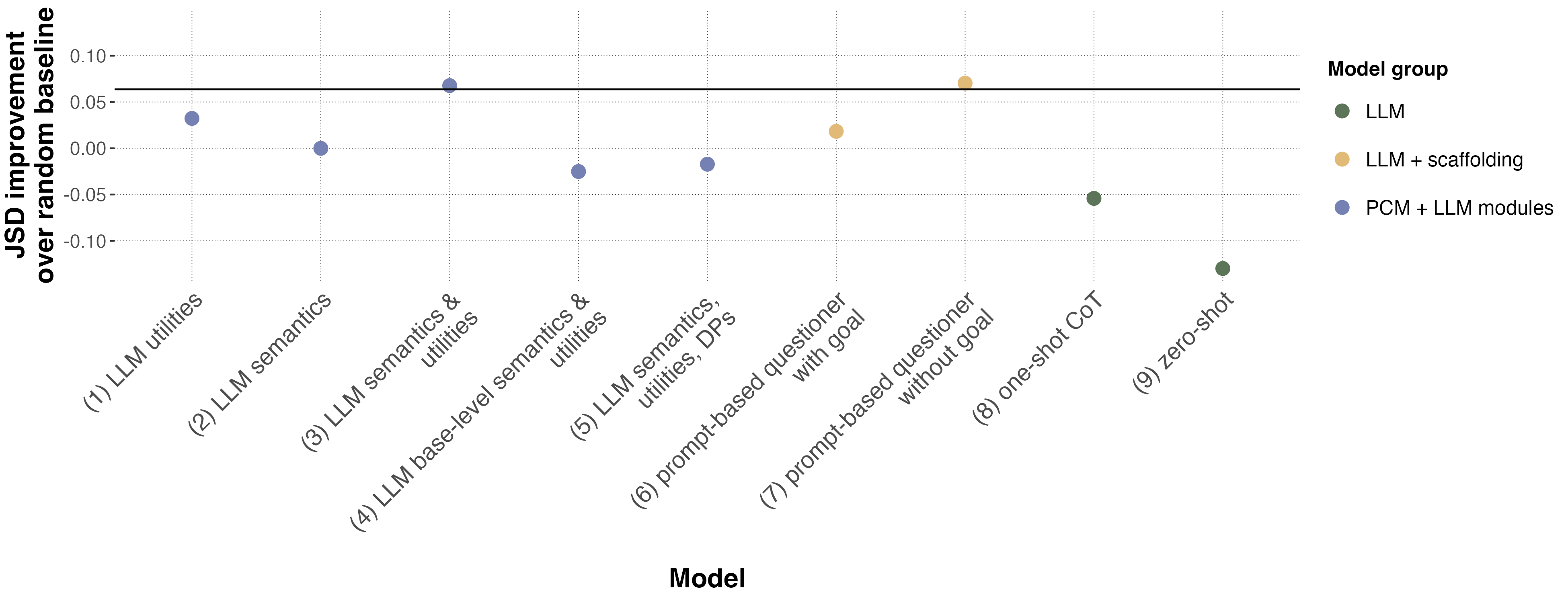}
	\caption{\label{fig:qwen-jsd}Improvement of the fit to human data of a model with an open-source Qwen-2.5-32B-Instruct backbone over a uniform response distribution baseline (higher is better). The horizontal line indicates the performance of the symbolic probabilistic model. The points indicate averages over simulations.}
\end{figure*}
\begin{figure*}[t]
	\centering
	\includegraphics[width=0.95\textwidth]{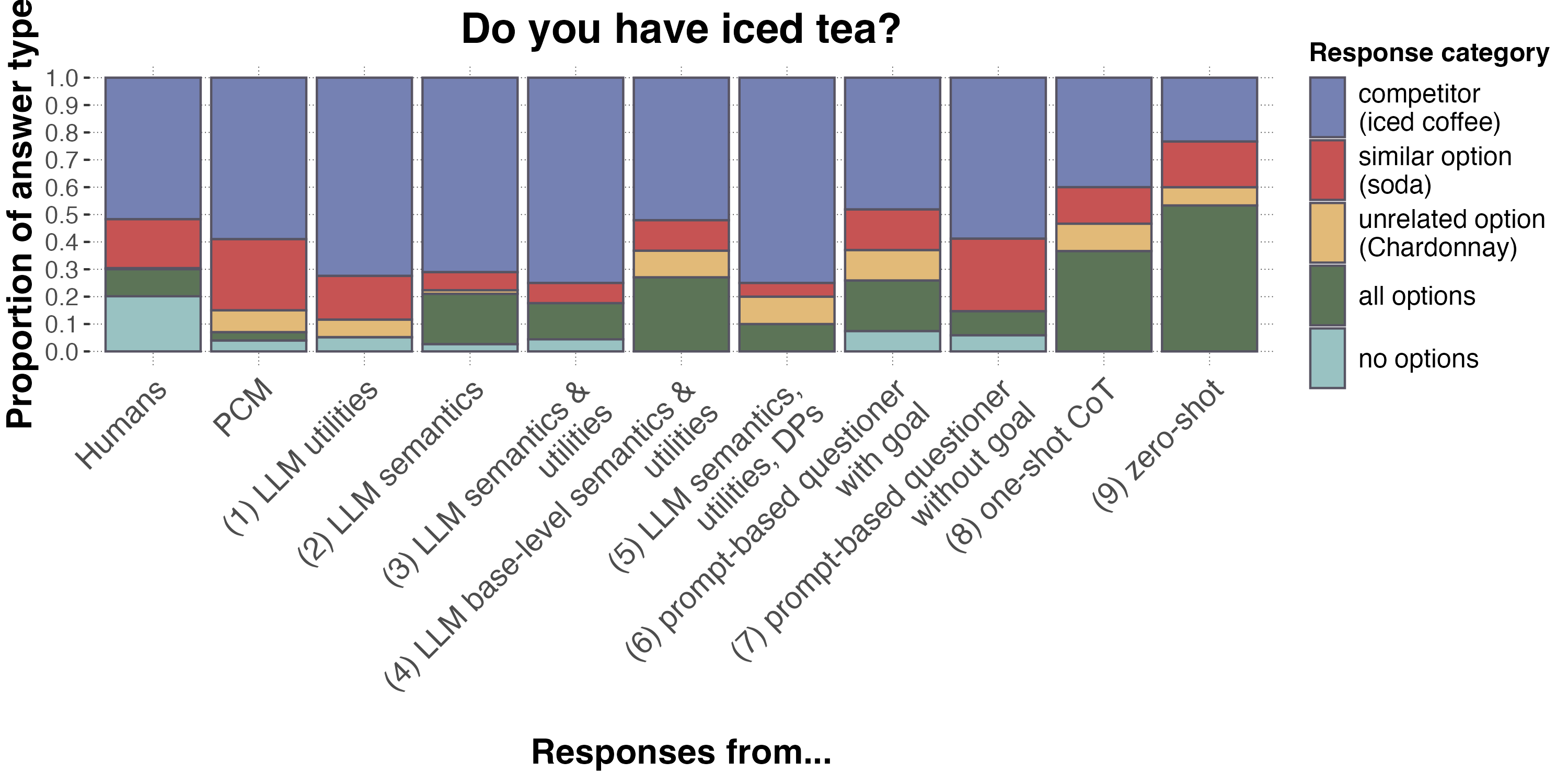}
	\caption{\label{fig:qwen-props}Proportions of different response categories predicted by Qwen-2.5-32B-Instruct used in different models (1--7), and with different prompting strategies (8--9).}
\end{figure*}

\begin{table*}[]
\centering
\begin{tabular}{l|l|l|l|l}
               & 1-shot CoT & 1-shot example & 1-shot explanation & 0-shot \\ \hline
Qwen-2.5-32B-Instruct     & 0.21   & 0.15          & 0.25         & 0.28         \\ \hline
Qwen-2.5-14B-Instruct  & 0.16   & 0.24         & 0.22         & 0.39        \\ \hline
Qwen-2.5-7B-Instruct & 0.33      & 0.19          & 0.50         & 0.17         \\ 

\end{tabular}
\caption{\label{tab:qwen-jsd}Jensen-Shannon divergence between human response proportions and the proportions of different response categories predicted by Qwen models of different sizes under various prompting (lower is better).}
\end{table*}
Additionally to the main experiments performed with \texttt{GPT-4o-mini}, we ran all experiments with an open-source LLM --- \texttt{Qwen-2.5-32B-Instruct} \cite{qwen2.5}, providing insights about advantages and open questions for our neuro-symbolic modeling framework when it is based on LLMs that can be run locally.

The experimental settings were the same as reported in~\ref{sec:procedure}.
Quantitative results comparing the predictions of the different models to human results in terms of JSD improvement over a random baseline $\Delta$, introduced in~\ref{sec:results}, are shown in Figure~\ref{fig:qwen-jsd}.
The results indicate that some models with LLM evaluators (i.e., semantics and utility evaluators, models (1) and (3)) perform on par with the models based on a powerful closed-source LLM, as well as close to the original probabilistic model. 
The high correlation between DP utilities predicted by Qwen and human results (Figure~\ref{fig:llm-utils}, right) corroborates that such evaluations can also be reliably elicited from an open-source model. 
Similarly to GPT-based models, the performance of the utility evaluator was more robust than for the literal semantic evaluators, as indicated by the better fit to human data for model (1).
However, for model (2) and for models introducing a proposer (models (4)--(5)) the fit of the models decreased. 
Manual evaluations of the single modules in these models indicated that, qualitatively, the generated evaluations and proposals were adequate for the respective modules. However, this LLM struggled more to follow formatting instructions, so that processing the proposals for passing them to the neural evaluator modules was more brittle. Simulation runs which resulted in unrecoverable parsing errors were excluded form analysis.\footnote{For this reason, no results of the full neuro-symbolic model are reported.}
Models which use a Qwen-based prompted questioner module ((6)--(7)) improved the fit to human data over the random baseline, although the role of conditioning the questioner prompt on the goal was opposite to the GPT-based models. 

Qualitative results comparing the proportions of different response types under different models are shown in Figure~\ref{fig:qwen-props}.  
The qualitative patterns suggest that Qwen-based models preferred responses mentioning a relevant alternative (i.e., competitor responses) over no options or exhaustive responses. LLM-only predictions, both in the one-shot chain-of-thought and the zero-shot prompting conditions, on the other hand, showed a larger proportion of exhaustive responses.  We also report the JSD values for predictions from different sizes of Qwen under different prompting strategies from \citet{tsvilodub2023overinformative} and human results in Table~\ref{tab:qwen-jsd}. These results suggest variation in the effectiveness of such prompting for different model sizes. For the two larger models, prompts that verbalize the PCM improve results over zero-shot prompting, although for the 32B model, ablated prompts further improve the fit to human data, suggesting substantial variation of human-likeness of the predictions when using only neural modules.

In sum, most neuro-symbolic Qwen-based models scaffolded with the PCM showed a better fit to human data than the random baseline, while the predictions of the LLM alone, even under one-shot chain-of-thought prompting, showed worse fit than the baseline. Additionally, given the open availability of the LLM, light-weight fine-tuning for better formatting instruction-following might offer a promising avenue for more robust neuro-symbolic modeling with open-source LLMs. Therefore, we can cautiously conclude that, given sufficient instruction-following capabilities for formatting, the neuro-symbolic framework might allow open-source LLMs to produce more human-like response patterns. 

\end{document}